\documentclass[conference]{IEEEtran}
\IEEEoverridecommandlockouts
\usepackage{cite}
\usepackage{amsmath,amssymb,amsfonts}
\usepackage{algorithmic}
\usepackage{graphicx}
\usepackage{multirow}
\usepackage{times}
\usepackage{latexsym}
\usepackage{url}
\usepackage{graphicx}
\usepackage[table,xcdraw]{xcolor}
\usepackage{colortbl}
\usepackage[normalem]{ulem}
\usepackage{amsmath}
\usepackage{ulem}
\usepackage{soul}
\usepackage{makecell}

\usepackage{multirow}
\usepackage{pgfplots}

\usepackage{comment}
\RequirePackage[inline,shortlabels]{enumitem}
\usepackage{textcomp}
\usepackage{xcolor}
\def\BibTeX{{\rm B\kern-.05em{\sc i\kern-.025em b}\kern-.08em
    T\kern-.1667em\lower.7ex\hbox{E}\kern-.125emX}}
 \pgfplotsset{compat=1.18} 
\begin{document}

\title{Retrieval and Augmentation of Domain Knowledge for Text-to-SQL Semantic Parsing\\
}


\author{
\IEEEauthorblockN{Manasi Patwardhan}
\IEEEauthorblockA{
\textit{TCS Research}\\
Pune, India \\
manasi.patwardhan@tcs.com}
\hspace{15em}
\and
\IEEEauthorblockN{Ayush Agarwal}
\IEEEauthorblockA{
\textit{TCS Research}\\
Pune, India \\
ayush.agarwal10@tcs.com}
\hspace{16em}
\and
\IEEEauthorblockN{Shabbirhussain Bhaisaheb}
\IEEEauthorblockA{
\textit{TCS Research}\\
Pune, India \\
shabbirhussainb@gmail.com}
\hspace{17em}
\and
\IEEEauthorblockN{Aseem Arora}
\IEEEauthorblockA{
\textit{TCS Research}\\
Pune, India \\
aseem\_1911mc02@iitp.ac.in}
\hspace{16em}
\and
\IEEEauthorblockN{Lovekesh Vig}
\IEEEauthorblockA{
\textit{TCS Research}\\
Delhi, India \\
lovekesh.vig@tcs.com}
\hspace{16em}
\and
\IEEEauthorblockN{Sunita Sarawagi}
\IEEEauthorblockA{
\textit{IIT Bombay}\\
Mumbai, India \\
sunita@iitb.ac.in}
\hspace{16em}
}

\maketitle

\begin{figure*}[!h]
\begin{center}
\includegraphics[width=\textwidth]
{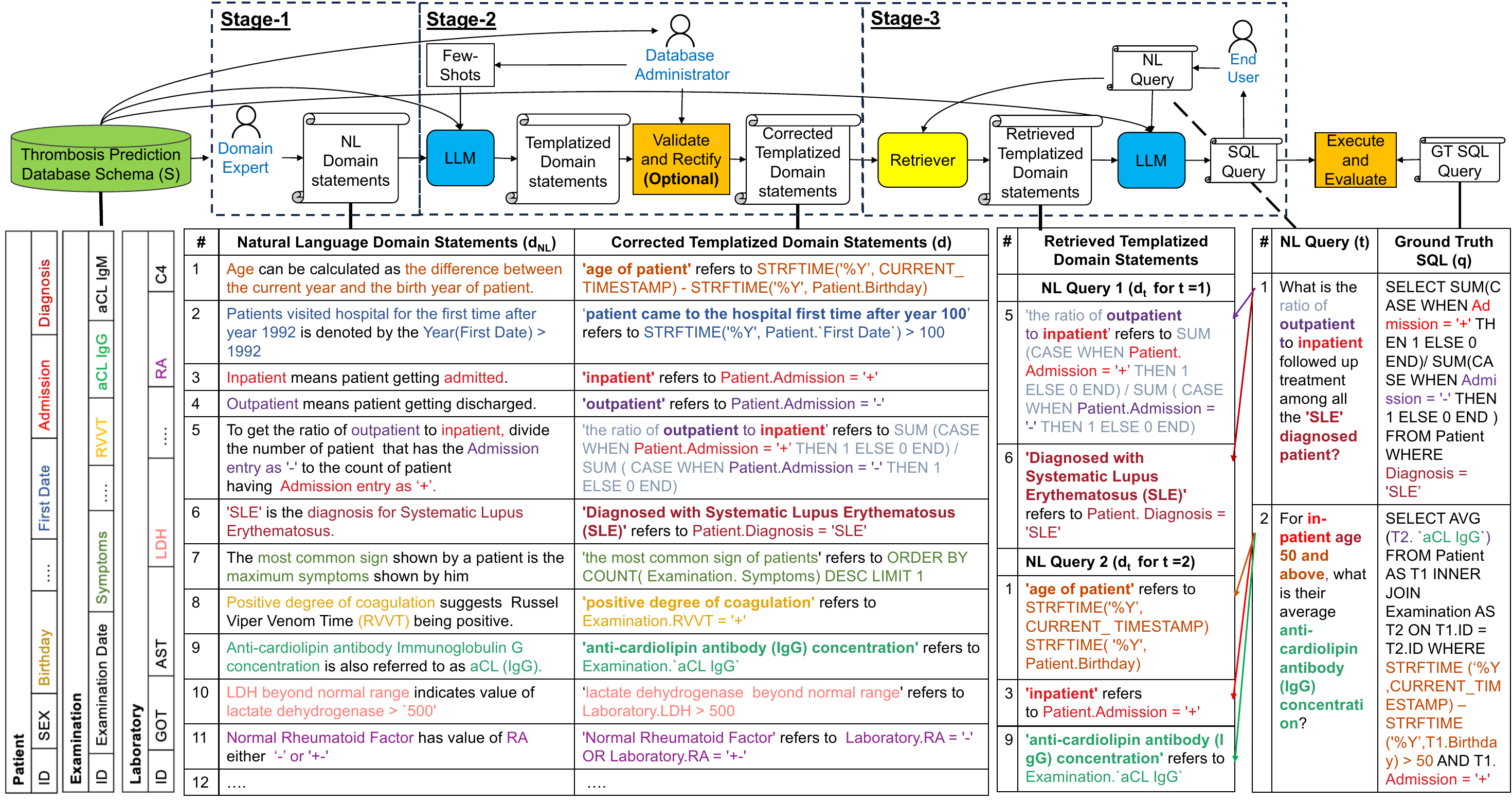}
\caption{ Framework for Enterprise NL-SQL Semantic Parsing. Mentions of same entities across DB Schema, NL and Structured Domain Statements, NL Query and Ground Truth SQL, are coded with the same color.
}
\label{fig:example}
\end{center} 
\end{figure*}

\begin{abstract}
The performance of Large Language Models (LLMs) for translating Natural Language (NL) queries into SQL varies significantly across databases (DBs). NL queries are often expressed using a domain specific vocabulary, and mapping these to the correct SQL requires an understanding of the embedded domain expressions, their relationship to the DB schema structure. Existing benchmarks rely on unrealistic, ad-hoc query specific textual hints for expressing domain knowledge. In this paper, we propose a systematic framework for associating structured domain statements at the database level. We present retrieval of relevant structured domain statements given a user query using sub-string level match.  We evaluate on eleven realistic DB schemas covering diverse domains across five open-source and proprietary LLMs and demonstrate that (1) DB level structured domain statements are more practical and accurate than existing ad-hoc query specific textual domain statements, and (2) Our sub-string match based retrieval of relevant domain statements provides significantly higher accuracy than other retrieval approaches.
\end{abstract}

\begin{IEEEkeywords}
Knowledge Augmented Text to SQL, Efficient Knowledge Retrieval,
Question Answering on Databases\end{IEEEkeywords}

\section{Introduction}
\begin{table*}[t]
    \centering
    \caption{Four Categories of Domain Statements illustrating types of domain understanding required for DBs.}
    \footnotesize
    \resizebox{\textwidth}{!}{
    \begin{tabular}{|l|l|l|l|}
    \hline
    \textbf{Category} &
    \textbf{Database} &
    \textbf{Natural Language Domain Statement} &
    \textbf{Structured Domain Statement} \\ 
    \hline
    
    \multirow{2}{*}{
    \begin{tabular}[c]{@{}l@{}}
    \textbf{Description } \\
    \textbf{of Cryptic} \\
    \textbf{Schema Names }
    \end{tabular}
    }
    & 
    Financial
    &
    \begin{tabular}[c]{@{}l@{}}
    A12 indicates the unemployment \\
    ratio for year 1995.  
    \end{tabular}
    &
    \begin{tabular}[c]{@{}l@{}}
    `unemployment ratio of year 1995' \\
    refers to district.A12
    \end{tabular}
    \\
    \cline{2-4}
    & 
    Formula 1
    &
    \begin{tabular}[c]{@{}l@{}}
    Location coordinates of the circuit of the race \\
    are given be a pair of latitude and longitude
    \end{tabular}
    &
    \begin{tabular}[c]{@{}l@{}}
    `location coordinates of the circuit of the race'\\
    refers to circuits.lat, circuits.lng 
    \end{tabular}
    \\
    \hline
    \multirow{2}{*}{
     \begin{tabular}[c]{@{}l@{}}
    \textbf{Description} \\
    \textbf{Categorical} \\
    \textbf{Column Values}
    \end{tabular}
    }
    & 
    \begin{tabular}[c]{@{}l@{}}
    Calif. Schools
    \end{tabular}
    &
    \begin{tabular}[c]{@{}l@{}}
    Amador is a school county 
    in California
    \end{tabular}
    &
    \begin{tabular}[c]{@{}l@{}}
    `Amador' refers to 
    schools.County = `Amador'
    \end{tabular}
    \\
    \cline{2-4}
    & 
    \begin{tabular}[c]{@{}l@{}}
   Toxicology
    \end{tabular}
    &
    \begin{tabular}[c]{@{}l@{}}
    Non-carcinogenic molecules are labelled as ‘-’
    \end{tabular}
    &
    \begin{tabular}[c]{@{}l@{}}
    `non-carcinogenic molecules' refers to \\molecule.label = ‘-’
    \end{tabular}
    \\
    \hline
    \multirow{3}{*}{
    \begin{tabular}[c]{@{}l@{}}
    \textbf{Handling } \\
    \textbf{Format} \\
    \textbf{Mismatches}
    \end{tabular}
    }
    & 
    \begin{tabular}[c]{@{}l@{}}
    Debit Card\\
    Specializing
    \end{tabular}
    &
    \begin{tabular}[c]{@{}l@{}}
    September 2013 is represented as\\
    the date format year month = `201309'
    \end{tabular}
    &
    \begin{tabular}[c]{@{}l@{}}
    `September 1000' refers to 
    yearmonth.Date \\= `100009'
    \end{tabular}
    \\
    \cline{2-4}
    & 
    \begin{tabular}[c]{@{}l@{}}
    Codebase\\
    Community
    \end{tabular}
    &
    \begin{tabular}[c]{@{}l@{}}
    Last accessed by user after the date 2014/9/1 \\
    means the Last Access Date $>$`2014-09-01'
    \end{tabular}
    &
    \begin{tabular}[c]{@{}l@{}}
    `last accessed after date 1000/10/10' refers to \\ date(users.LastAccessDate) $>$`1000-10-10'
    \end{tabular}
    \\
    \hline
    \multirow{2}{*}{
    \begin{tabular}[c]{@{}l@{}}
    \textbf{Converting} \\
    \textbf{NL Expression}\\
   \textbf{to Formula} 
    \end{tabular}
    }
    & 
    \begin{tabular}[c]{@{}l@{}}
    European\\
    Football2
    \end{tabular}
    &
    \begin{tabular}[c]{@{}l@{}}
    Highest potential score is an attribute player \\
    calculated by taking taking maximum of potential
    \end{tabular}
    &
    \begin{tabular}[c]{@{}l@{}}
    `highest potential score' refers to  ORDER \\ 
    BY Player\_Attributes.potential DESC LIMIT 1
    \end{tabular}
    \\
    \cline{2-4}
    & 
    \begin{tabular}[c]{@{}l@{}}
    Thrombosis \\ 
    Prediction
    \end{tabular}
    &
    \begin{tabular}[c]{@{}l@{}}
    Normal level of complement 3 \\
    suggests value of C3 $>$ 35
    \end{tabular}
    &
    \begin{tabular}[c]{@{}l@{}}
    ‘normal level of complement 3’ \\
    refers to Laboratory.C3 $>$ 35
    \end{tabular}
    \\
    \hline
    \end{tabular}
    }
    \label{tab:Examples}
\end{table*}
The impressive natural language understanding and code generation capabilities of modern LLMs has led to significantly improved performance on NL-SQL semantic parsing \cite{Liu2024ASO,Li2024TheDO}. 
However, their accuracy 
varies widely with the database queried \cite{Lan2023UNITEAU}. DBs in WikiSQL\cite{Zhong2017Seq2SQLGS} or Spider\cite{yu2018spider}  contain semantically meaningful table/column names and cell values making it easier for LLMs to accurately link domain expressions in the NL query with the DB schema/cell elements. For more challenging DBs with semantically meaningless entities such as the DBs in KaggleDBQA\cite{lee-2021-kaggle-dbqa} or XSP\cite{suhr-etal-2020-exploring}, average performance is significantly worse\footnote{KaggleDBQA: \url{https://paperswithcode.com/sota/text-to-sql-on-kaggledbqa}}. This variation in performance often stems from the domain-specific vocabulary used to frame the NL queries and providing LLMs with additional context information in the form of relevant Domain Statements (DSs) may help address this problem.

The large performance gap on the recent BirdSQL benchmark \cite{li2024can} between approaches with and without the use of NL query-specific oracle domain knowledge\footnote{\label{fn:bird}BirdSQL: \url{https://bird-bench.github.io}} indicates an indispensable need for domain knowledge augmentation for NL query resolution. Mapping the NL queries to their correct SQL  requires an understanding of (i) the embedded domain expressions, and (ii) their relationship to the schema elements and structure of the DB. BirdSQL fails to suitably capture this dichotomy, and relies on unrealistic, ad-hoc query-specific textual hints for expressing domain knowledge. In this paper, we propose a systematic framework for soliciting DSs at the DB level, as opposed to the NL query level, in a format that optimizes LLM performance on downstream real-world enterprise semantic parsing.

Figure \ref{fig:example} presents the outline of the framework. Over the life-cycle of an enterprise DB,  DSs are solicited from  domain experts, such as bankers and doctors, incrementally as and when they identify domain expressions for which additional information is needed (\S\ref{sec:sol}). The DSs are expected to be in NL form $d_{NL}$, as domain experts may not be familiar with the DB structure or SQL, but would likely be adept at providing the information required to understand domain expressions in NL form.   
For larger enterprise DBs, as the DSs get accumulated, they may be too numerous for LLMs with smaller context to accommodate for LLM driven NL-to-SQL semantic parsing. Also, including irrelevant statements in the context can confuse the LLM \cite{Shi2023LargeLM}.  Thus, there is a need to retrieve relevant DSs for the NL query. Aiming to facilitate the retrieval and downstream semantic parsing, these $d_{NL}$,  are to be translated into a unified format that lends itself to optimized retrieval and LLM driven semantic parsing. For this, we define our own structured format for domain expressions ($d$) comprising of NL expressions  mapped to SQL code snippets. Whenever a $d_{NL}$, is received from the domain expert, it is translated to $d$ via LLM based few shot in-context learning and any errors in the generated $d$ may optionally be corrected by the Database Administrator (DBA) (\S\ref{sec:tras}). The end user application then retrieves the appropriate structured DSs for the  NL query (\S\ref{sec:ret}). Thus, the process of DS structuring is  automated and is executed when domain repositories get updated. Also, the retrieval of DSs is automated and is triggered when an end-user poses a query on the DB.

Figure \ref{fig:example} illustrates  example 
NL DSs $d_{NL}$, solicited for a `Thrombosis Prediction' DB 
. They provide information about DB column (2,9,10), cell values (3,4,6,8,11) and derived entities (1,5,7). As observed, each NL-query ($t$) may require multiple DSs to parse (DS 5, 6 for query 1 and 1, 3, 9 for query 2) and each relevant $d_{NL}$, has a semantic match with only a partial NL-query. 
Noting that a $d_{NL}$, relevant to a query should act as a knowledge `link' between the NL query and its SQL counterpart, we design structured DSs ($d$) with two components. The `text' component helps the LLM interpret the NL query, and the `SQL snippet' provides hints for generating the desired SQL query. For example, for the structured DS 2 in Figure \ref{fig:example}, \textit{Patient came to the hospital first time after year 1992} depicts the `text' and \textit{$STRFTIME$(`$\%Y$', Patient.`First Date') $>$ 1992} depicts the `SQL snippet', with both parts connected by the `refers to' phrase. 
We further define a strategy for semantic matching of query sub-strings with the `text' component of $d$ to rank and retrieve top-K DSs.  Along with the DB schema, these retrieved DSs are included in the LLM context for SQL generation.

Current State-of-the-art NL-to-SQL approaches \cite{Gao2024APO, Pourreza2024CHASESQLMR, li2024codes, pourreza2024dts} for BirdSQL, utilize query specific oracle DSs provided in the dataset. Soliciting DSs from domain experts for each new query posed by the end-user on the DB in real-time is unrealistic. Hence, we release a more realistic version of this dataset where the DSs are solicited at the DB schema level, structured with clearly demarcated NL domain expression and SQL snippets (\S\ref{sec:dataset}). For each query posed on the DBs of the BirdSQL Dev set, we retrieve relevant structured DSs from the DB level repository using our retrieval technique. We demonstrate that, by augmenting these retrieved structured DSs in the LLM context we obtain higher accuracy for the downstream SQL generation task, as opposed  to augmenting the context with query specific oracle DSs available in the original BirdSQL.
Our main contributions are:

\begin{itemize}
\item We outline a framework for practical NL-to-SQL for enterprise DBs requiring domain knowledge for NL query understanding. As the part of the framework, we define a new structure for DS Structuring that helps to link an NL query to its equivalent SQL (\S\ref{sec:tras}). 
\item We extend the BirdSQL Dev set to benchmark solutions utilising query-agnostic DSs for NL-SQL task (\S\ref{sec:dataset}). 
\item We devise a technique to retrieve structured DSs relevant to an NL query, based on a semantic match of query sub-strings and the textual `link' of structured DSs (\S\ref{sec:ret}).
\item We perform exhaustive experiments to evaluate our pipeline demonstrating the value of different components including: (i) DS inclusion, (ii) structuring of NL DSs (iii) DBA correction and (iv) sub-string based retrieval, for eventual SQL generation (\S\ref{sec:results}).  Our approach outperforms approaches with none or all DSs as well as retrieved DSs with other baselines, across LLMs, highlighting the impact of the robust retrieval mechanisms. 
\end{itemize}
\vspace{-1mm}

\begin{figure*}[!h]
\begin{center}
\includegraphics[width=\textwidth]
{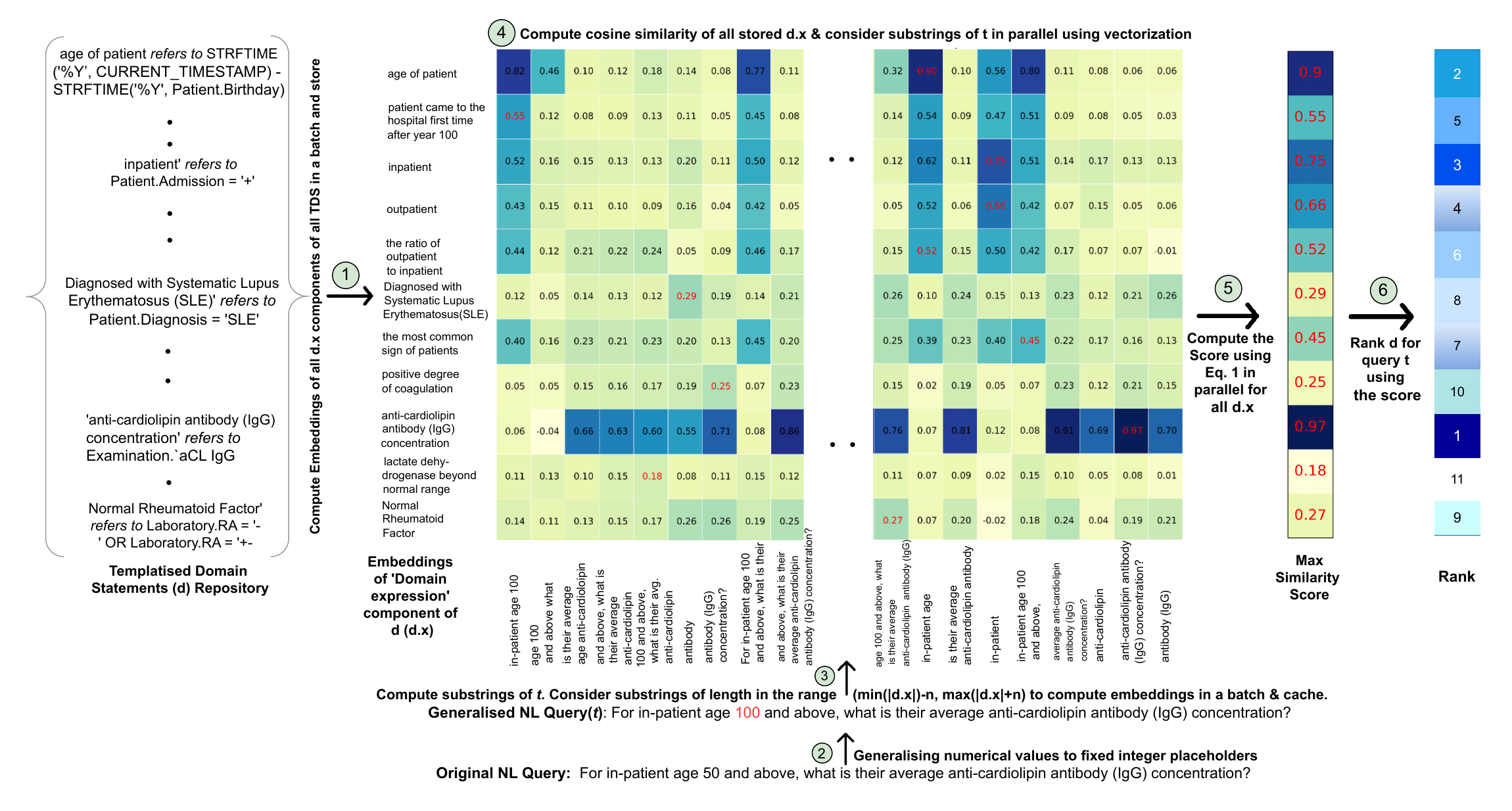}
\caption{ Illustrative Example for Sub-String based Ranking (SbR) Mechanism
}
\label{fig:example1}
\end{center} 
\end{figure*}

\begin{table}[!t]
\centering
\small
\caption{LLM Prompt for Domain Statements translation}
\label{tab:TDSprompt}
\resizebox{\columnwidth}{!}{
\begin{tabular}{l}
\hline
\textbf{System Prompt:} \\ 
\hline
\begin{tabular}{l} 
You are a database administrator. You have been given a few examples of \\domain statement and
their corresponding templatized DSs. The DSs are \\ for the database $<$DB name$>$ whose schema is provided below:\\
CREATE TABLE Patient ( \\  
ID INTEGER  PRIMARY KEY,  -- identification of the patient;\\
SEX TEXT , -- Sex; \\
Birthday DATE, --Format Year-Month-Date;\\
Diagnosis TEXT  -- disease names;) 
...\\
 \hline
\end{tabular} \\ 
\textbf{ User Prompt:} \\ \hline
\begin{tabular}{l} 
For the given domain statement, you need to create a templatized statement.\\
Provide the statement in one sentence only. Generate the templatized \\
domain statement in the following form:`Relevant text component' refers \\
to `corresponding SQL syntax' for each sample.\\
\#\#\# Few Shot Exemplars\\
Example:\\
Domain Statement: female is also represented as `F'\\
Templatized Domain Statement:`female' refers to Patient.SEX = `F'\\
Example:\\
Domain Statement: patient born between Year 1930 to 1940 represents \\Birthday BETWEEN the dates `1930-12-31' AND `1940-01-01'\\
Templatized Domain Statement: `patient born between year 100 and 100' \\  refers to STRFTIME(`\%Y', Patient.Birthday) BETWEEN `100' AND `100'\\ 
...\\
\hline
\end{tabular}
\end{tabular}
}
\end{table}

\begin{table}[!t]
\centering
\caption{LLM Prompt for SQL Generation }
\label{tab:SQLprompt}
\small
\resizebox{\columnwidth}{!}{%
\begin{tabular}{l}
\hline
\textbf{ System Prompt:} \\ \hline
\begin{tabular}{l} 
You are a database administrator. You have designed the following\\
database for $<$DB name$>$ whose schema is represented as:\\
CREATE TABLE Examination (\\
ID INTEGER PRIMARY KEY, -- identification of the patient\\
`Examination Date` DATE, --Format Year-Month-Date;\\
`aCL IgG` REAL, -- anti-Cardiolipin antibody (IgG) concentration\\
ANA INTEGER , -- anti-nucleus antibody concentration\\   
Diagnosis TEXT ,  -- disease names\\
Thrombosis INTEGER , --degree of thrombosis
); ...
\end{tabular} \\
\hline 
\textbf{ User Prompt:} \\ 
\hline
\begin{tabular}{l} 
Query: $<$User Query$>$\\
Domain Knowledge statements, some of which might or \\
might not be useful to generate the query:$<$A list of DSs$>$ \\ 
Generate a single SQL in SQLite format for the above query. Do not \\include any extra text, tag or information other than the SQL query itself.\\
SQL:
\end{tabular}  \\
\hline
\end{tabular}
}
\end{table}

\section{Related Works}

Existing approaches for  NL-to-SQL semantic parsing  \cite{rai2023improving, li2023graphix, gao2023text, pourreza2024din} assume DBs with semantically meaningful table, column names such as Spider\cite{yu2018spider}, WikiSQL \cite{zhongSeq2SQL2017}, or GeoQuery \cite{Zelle:1996:LPD:1864519.1864543}.
Results on these DBs often fail to translate to enterprise DBs with semantically opaque schema elements and missing domain information.  Synthetically created datasets incorporate domain knowledge (SpiderDK \cite{Gan2021ExploringUL}), replace schema-related words by synonyms (Spider-Syn \cite{Gan2021TowardsRO}) or paraphrase words (DBPal \cite{Utama2018AnEN}) in queries, allowing for increased difficulty for query understanding. However, for enterprise DBs, the required domain knowledge can be beyond column, value descriptions, including knowledge of entities derived with mathematical or string operations.

The BirdSQL \cite{li2024can} dataset contains realistic DBs from multiple domains with entities requiring additional information for query understanding, provided in the form of the oracle query specific domain knowledge (evidences). 
Queries in recently published Spider 2.0 dataset \cite{lei2024spider} have a similar requirement. However, we observe that the dataset does not explicitly provide such domain knowledge and thus does not facilitate the simulation of our framework (\S\ref{sec:approach}).

 As opposed to existing few-shot retrieval techniques \cite{an2023skill, li2023resdsql,rubin-etal-2022-learning} , our focus is on retrieval of domain statements. 
 Existing approaches perform retrieval \cite{dou2023towards} or generation \cite{Hong2024KnowledgetoSQLES} of domain knowledge for the downstream text-to-SQL task. They assume the availability of parallel data in the form of a query and a DS pair to fine-tune the base model. As opposed to these, we assume no availability of such data for a given DB.  Moreover, in an enterprise setting, applicability of cross-domain DB knowledge is infeasible. This is also validated by our observation with distinct DBs within BirdSQL, where a match of domain statements across DBs is incidental and rare and possible only in case of standard public knowledge (genders, date formats, etc). By definition the DSs are needed when DBs have their own traits. Hence, we assume an in-domain (single DB) setting where there would be availability of DB level domain knowledge, as opposed to cross-domain setting assumed in \cite{Hong2024KnowledgetoSQLES}. We find it surprising that the DS generator in \cite{Hong2024KnowledgetoSQLES} would generalize across schema. Subsequently, we also observe our gains (15.3\%) are much more over \cite{Hong2024KnowledgetoSQLES} (5.7\%), primarily because we lean on DSs provided by domain experts, as opposed to generating them.

\section{Framework for NL-SQL Semantic Parsing}\label{sec:approach}
In our proposed framework,  a domain expert can provide domain statements $d_{NL}$ in natural language  at any time during the usage of a database (\S~\ref{sec:sol}).   We translate each $d_{NL}$ into a structured form $d$ that clearly demarcates the mapping between a natural language phrase and an SQL snippet (\S\ref{sec:tras}). The structured DS $d$ are stored and indexed as part of other meta data associated with the DB.
Any subsequent user query $t$, will first retrieve a subset of the relevant DSs $d_t \subseteq d$ (\S~\ref{sec:ret}) to augment for the final text-to-sql generation. 

\subsection{Natural Language Domain Statements}\label{sec:sol}
Enterprises invest in one-time manual effort to build domain knowledge for their long lived DBs for gains in their reliability and accuracy. Domain experts may not necessarily be fluent in SQL, and specify domain knowledge in NL statements. 
We identify four broad categories of such statements (Table \ref{tab:Examples}): (i) description of cryptic schema names, 
(Figure~\ref{fig:example} DS 9), (ii) description of string values of categorical columns
(Figure \ref{fig:example} DSs 3, 4 and 8), (iii) Handling format mismatches 
especially for  numerical and date values. 
(Figure \ref{fig:example} DS 2), and (iv) Converting NL expression to formula
for specifying domain-specific predicates such as age, ratio of in and out patients, etc
(Figure \ref{fig:example} DS 1, 5, 7, 10, 11). We believe that such DSs by definition are very specific and private to enterprise DBs and cannot be conjured from LLMs trained on public data. Thus their solicitation from domain experts is unavoidable. 

\begin{table}[h]
\centering
\caption{BirdSQL \cite{li2024can} Dev Split DBs. \textbf{DS}: Union of Domain Statements (Evidences) of IN Set Questions}
\label{tab:db_info}
\footnotesize
\begin{tabular}{lcccc}
\hline
\textbf{Database}                  & \textbf{\#Tables} & \textbf{\#Columns} & \textbf{\#Questions} & \textbf{\#DS} \\ \hline
Thrombosis Pred.    &  3               &   72             & 163        & 188         \\ 
Calif. School        & 3               &    89            & 89         & 52          \\ 
Card Games               & 6              &     168          & 191        & 185         \\ 
Debit Card Spec. & 5             &    21            & 64         & 33          \\ 
Toxicology               & 4               &    11            & 145        & 300         \\ 
Financial                 & 8               &     49           & 106        & 66          \\ 
Codebase Comm.      & 8               &     71           & 186        & 136         \\ 
Euro. Football2      & 7              &    199            & 129        & 124         \\ 
Formula 1                 & 13              &    94           & 175        & 102         \\ 
Student Club              & 8               &    48            & 158        & 127         \\ 
Superhero                & 10               &    31           & 129        & 130         \\ \hline
Average               & 6.8               &    77.5           & 139.5        & 131.2         \\ \hline
\end{tabular}%

\end{table}

\subsection{Domain Statement Structuring}\label{sec:tras}
 We translate $d_{NL}$ into a structured form $d$ that differs from the original  $d_{NL}$ in two ways: (1) $d$ is of the form "$d.x$ refers to $d.q$" where  $d.x$ denotes a NL domain expression, and $d.q$ denotes a SQL snippet to which $d.x$ maps. 
Such a representation of $d$ allows to better match domain specific constructs in the user query, and lookup the attached SQL snippet.  (2) Constants such as names and dates in  $d_{NL}$ are converted to a fixed canonical form to allow better match with user queries. We expect the structuring to bring the NL DSs into a unified format, leading to more generalizable downstream retrieval. Figure~\ref{fig:example} and the fourth column of Table~\ref{tab:Examples}  provide examples of structured DSs for all the four categories, discussed in section \ref{sec:sol}. 

We harness LLMs for translating  $\{d_{NL}\}$ to $\{d\}$. For each DB, the DBA defines few-shot exemplars consisting of pairs of $<d_{NL},d>$ covering all the FOUR categories ($\sim$2 per category).
Given a DB schema $S$, the few-shots $\{<d_{NL}, d>\}$ provided by the DBA and $d_{NL}$ provided by the domain expert, with a proper instruction, the LLM generates a $d$ for each $d_{NL}$ (Table \ref{tab:TDSprompt}).  We compare $d$ generated by the LLM  with the ones manually translated by DBA, for three DBs (Thrombosis Prediction, California Schools and Card Games). We observe to have on an average 77.67\% lexical (word)  and 88.82\% semantic overlap, computed using BERT embeddings \cite{reimers2019sentence}. This demonstrates the ability of the LLM to generate $d$ from $d_{NL}$ and also  the amount of noise in the LLM generated $d$.
To ensure correctness, given $S$, $d_{NL}$, and $d$ generated by the LLM, the DBA \emph{optionally} validates $d$ and rectifies errors, if any.  For this process the DBA is NOT provided with any information about NL-SQL query pairs to which the DSs are aligned. 
Once the $d$ are created, we compute and index the embeddings of their text components ($d.x$).

\subsection{Structured Domain Statement Retrieval}\label{sec:ret}
Given a user query $t$, an LLM generates an SQL $q$, when instructed with a prompt comprising of the query $t$, the DB schema $S$ including column name and value descriptions, augmented with relevant DSs (Table \ref{tab:SQLprompt}). 
We devise a retriever $R$, which retrieves a relevant subset $\{\hat{d_t}\} \subseteq \{d\}$ of the DSs given $t$ using $d.x$.  A standard method of retrieval augmentation is to match the query embedding  $t$ to the  embeddings of each stored $d$. This approach is inadequate for our case, as a $d$ may be relevant only to a part of a query, and multiple DSs may map to different parts of $t$.  For example, different parts of the user query 2 in Figure~\ref{fig:example} matches the `text' ($d.x$) of three relevant DSs. 
This motivates the need for our query decomposition-based retrieval procedure, described  below.

Given a user query  $t$, we convert it into a generic form by replacing  numerical values and dates to the same fixed integer placeholder value used during DS Structuring for better matching.  The next step is to compute the embeddings of the contiguous sub-strings of the query $t$  and use cosine similarity (SIM) to compute their matches with the stored embeddings of $d$.
The complexity of the embedding and similarity computation increases exponentially  with the number of sub-strings of $t$ and thus is of the order of $2^{|t|}$, where $|t|$ is the query length. We observe that for a $d$, only those contiguous sub-stings of $t$, having lengths in the range of $(\lfloor |d.x| - n \rfloor , \lceil |d.x| + n \rceil)$, $n$ being the threshold, yields the best match.  Considering only these subset of  sub-strings, the complexity is the order of $2^{|d.x|}$. $|d.x|$  (average 4.28 words for all DBs in the dataset)  is typically much smaller (4 times) than $|t|$ (average 14.55 words), leading to efficiency gains for computations of embeddings as well as matching. 
For a new DB, we assume the availability of a few NL-SQL pairs to tune the threshold $n$. We set it to a value which yields maximum execution accuracy for downstream SQL generation for these available NLs, when augmented with the retrieved $d_t$s.
For each new query $t$ posed on a db,  we pre-compute the embeddings of only those contiguous sub-strings within the range of ($\lfloor min(|d.x|) -n \rfloor$, $\lceil max(|d.x|) + n \rceil$) using batch processing and cache them. Here $min(|d.x|)$ and $max(|d.x|)$ are the minimum and maximum  lengths of $d.x$ components of all $d$s available in the repository for that DB. We use the cached embeddings of these query sub-strings to compute their similarity with all $d$s in the repository in parallel using vectorization. 
We define the similarity between $t$, and a $d$ as follows:
Let $t_1, ...., t_m$ denote the ordered list of the words in the query. As depicted in Equation \ref{eq:subseq}, the score between a query $t$ and a structured DS $d$, is computed as the maximum of the similarity of $d.x$ with the considered sub-strings of $t$, determining it's best match. 
\begin{equation}
\label{eq:subseq}
Score(t, d) = \max_{ |d.x| - n  \le i \le j \le  |d.x| + n }\text{SIM}(d.x, [t_i,t_j])
\end{equation}
For a query, we  rank  the $d$s using this scoring mechanism and augment top-K to  the prompt for SQL generation. We reduce the time taken to perform retrieval for an average length query from few minutes to few milliseconds. For large DS repositories we envision retrieval of  candidate DSs by using ColBERT \cite{santhanam-etal-2022-colbertv2} style index, followed by re-ranking with our method. An illustration of this efficient sub-string based retrieval method is provided (Figure \ref{fig:example1}). Following this, we perform SQL generation via LLM with the retrieved relevant top-k DS in context (Table \ref{tab:SQLprompt}).

\begin{table*}[!ht]
\centering
\footnotesize
\caption{Examples of Missing and Erroneous Oracle Domain Statements in BirdSQL indicated by \textcolor{red}{Red} 
}
\resizebox{\textwidth}{!}{%
\begin{tabular}{|llll|}
\hline
\multicolumn{1}{|l|}{\textbf{\begin{tabular}[c]{@{}l@{}}\#\end{tabular}}} &
  \multicolumn{1}{l|}{\textbf{Database}} &
  \multicolumn{1}{l|}{\textbf{Question}} &
  \textbf{Natural Language Domain Statements} \\ \hline
\multicolumn{4}{|c|}{\textbf{Missing Domain Statements}} \\ \hline
\multicolumn{1}{|l|}{1.} &
  \multicolumn{1}{l|}{\begin{tabular}[c]{@{}l@{}}Euro. Football2\end{tabular}} &
  \multicolumn{1}{l|}{\begin{tabular}[c]{@{}l@{}}What's the heading accuracy of Ariel Borysiuk?\end{tabular}} &
  \begin{tabular}[c]{@{}l@{}}\textcolor{red}{Ariel Borysiuk is player name}\end{tabular} \\ \hline
\multicolumn{1}{|l|}{2.} &
  \multicolumn{1}{l|}{Formula 1} &
  \multicolumn{1}{l|}{\begin{tabular}[c]{@{}l@{}}Name the top 3 drivers and the points they scored \\in the 2017 Chinese Grand Prix.\end{tabular}} &
  \begin{tabular}[c]{@{}l@{}}\textcolor{red}{Name of the drivers consists of their forename} \\ \textcolor{red}{and surname.}\end{tabular} \\ \hline
\multicolumn{1}{|l|}{3.} &
  \multicolumn{1}{l|}{\begin{tabular}[c]{@{}l@{}}Debit Card Spec.\end{tabular}} &
  \multicolumn{1}{l|}{\begin{tabular}[c]{@{}l@{}}Which year recorded the most gas use paid in EUR?\end{tabular}} &
 \begin{tabular}[c]{@{}l@{}}\textcolor{red}{Extract Year from Date by considering last 4 digits}\end{tabular} \\ \hline
\multicolumn{4}{|c|}{\textbf{Erroneous Domain Statements}} \\ \hline
\multicolumn{1}{|l|}{4.} &
  \multicolumn{1}{l|}{\begin{tabular}[c]{@{}l@{}}Thrombosis \\ Prediction\end{tabular}} &
  \multicolumn{1}{l|}{\begin{tabular}[c]{@{}l@{}}What was the age of the youngest patient when they \\initially arrived at the hospital?\end{tabular}} &
  \begin{tabular}[c]{@{}l@{}}the youngest patient is the one which has \\\textcolor{red}{MIN}(YEAR(Birthday))\end{tabular} \\ \hline
\multicolumn{1}{|l|}{5.} &
  \multicolumn{1}{l|}{\begin{tabular}[c]{@{}l@{}}Card Games\end{tabular}} &
  \multicolumn{1}{l|}{\begin{tabular}[c]{@{}l@{}}What percentage of cards without power are in French?\end{tabular}} &
  \begin{tabular}[c]{@{}l@{}}cards without power indicates value of \textcolor{red}{power = `*'}\end{tabular} \\ \hline
 
 
\multicolumn{1}{|l|}{6.} &
  \multicolumn{1}{l|}{\begin{tabular}[c]{@{}l@{}}California \\Schools\end{tabular}} &
  \multicolumn{1}{l|}{\begin{tabular}[c]{@{}l@{}}What is the phone number of the school  that has the highest \\number of test takers with an SAT score of over 1500?\end{tabular}} &
    \begin{tabular}[c]{@{}l@{}}\textcolor{red}{`False'}\end{tabular} \\ \hline
\end{tabular}
}
\label{tab:Error-DS}
\end{table*}

\section{Experimentation and Results}
\subsection{Dataset}\label{sec:dataset} 
We consider the BirdSQL Dev set, which is the standard dataset used for NL-to-SQL task. The DBs are from real platforms such as Kaggle, Relation.vit, etc, spanning over diverse  domains including medical, finance and sports. 
Each NL query posed on the DBs require domain understanding of one or more of the four categories ( \S\ref{sec:sol}), for which the dataset provides domain knowledge in terms of 1 to 3 evidences in natural language (Figure \ref{fig:example}, Table \ref{tab:Examples}). 
There are variations in the format in which the $d_{NL}$ are represented across the DBs. For example, the math formulae are specified as equations such as `\textit{Gap = highest average salary - lowest average salary}' (DB: Financial) or in text `\textit{Average score for all subjects can be computed by addition of AvgScrMath, AvgScrRead and AvgScrWrite}' (DB: California Schools).  Column values are described by various formats such as `\textit{`Chinese Simplified' is the language} (DB: Card Games) or `\textit{label = `+' means molecules are carcinogenic}' (DB: Toxicology). Our automated approach of LLM based Structuring shall translate  $d_{NL}$ in variety of different forms into the pre-specified structured form $d$,  demonstrating its  effectiveness (\S\ref{sec:tras}).
Additionally, the top performing approaches on BirdSQL\cite{Gao2024APO,Pourreza2024CHASESQLMR,li2024codes, pourreza2024dts, wang2023mac}  assume availability of these oracle query-specific DSs for SQL generation. In real-world enterprise settings, for each new query posed on a DB, such statements will not be available.  

 For the 11 DBs in Dev split of BirdSQL, we simulate the scenario of solicitation of DB level domain knowledge from domain experts. 
We form 50-50\%  random splits of the queries in each DB.
The queries in one of the splits are assumed to be the part of regular DB workload and the union of DSs of the queries in this split are assumed to be provided by the domain expert (\S\ref{sec:sol}),  to form the domain repository for retrieval. Henceforth, we refer to this split as the IN Set. Note, we do not assume the availability of the mapping between the DSs and queries. The DSs for the other split (OUT Set), are not included in this repository. We find that $\sim77\%$ of  queries in the OUT Set of all the DBs have no oracle DS overlap with the repository, forming an OUT-NO Set 
. 
Thus, we simulate the realistic scenario of domain repository construction with complete (IN Set), partial (OUT Set) and no (OUT-NO Set) knowledge of the queries that may get posed on the DB.  We expect an approach to be able to retrieve the right set-of DSs for the IN Set queries leading to good performance on the downstream SQL generation task, comparable to when the oracle DSs are in-context.  Whereas, we do not expect a comparable performance for the other sets. 
 Table \ref{tab:db_info} 
 provides the dataset statistics, which we release at
\footnote{\scriptsize\url{https://anonymous.4open.science/status/RAG-of-Domain-Knowledge-for-Text-to-SQL}}.

\subsection{Models and hyper-parameter Setting}\label{sec:hyper}
We use the following LLMs to generate SQL: \begin{enumerate*}
\item GPT-3.5-turbo
\footnote{\scriptsize\url{https://platform.openai.com/docs/models/gpt-3-5-turbo}}
\item Mixtral-8X7B
\footnote{\scriptsize\url{huggingface.co/mistralai/Mixtral-8x7B-Instruct-v0.1}} 
\item Llama3-70B
\footnote{\scriptsize\url{https://huggingface.co/meta-llama/Meta-Llama-3-70B}}
\item sqlcoder-8B
\footnote{\scriptsize\url{https://github.com/defog-ai/sqlcoder}}
\item Gemini-1.5 Flash
\footnote{\scriptsize\url{https://github.com/google-gemini/cookbook}}.
\end{enumerate*}
For reproducibility, we ensure deterministic outcomes by setting t = 0. We require better precision for IN Set queries with guaranteed availability of the required domain knowledge and better recall for OUT Set queries, which may have only partial domain knowledge available. Hence, for the retriever, we set lower value of K = 4 for the IN Set and higher value of K = 10 for the OUT Set.
We compute embeddings with best performing\footnote{\scriptsize \url{https://www.sbert.net/docs/pretrained_models.html}}  \textit{all-mpnet-base-v2} BERT based model\footnote{\scriptsize \url{https://huggingface.co/sentence-transformers/all-mpnet-base-v2}} \cite{reimers2019sentence}. We get the best performance of the sub-string based matching technique  at the threshold $n$ (\S\ref{sec:ret}) ranging from 2 to 4 for the DBs in the BirdSQL Dev split.

\subsection{ Evaluation Metric}\label{sec:eval}
We use `Execution Accuracy' of the generated SQLs with DSs in context as the evaluation metric, which is invariant to the syntactic variations in the generated and gold SQLs. 
Analyzing the dataset, we observe that there are 20.17\% queries with missing (examples 1-3) and 6.72\%  with erroneous (examples 4-6) oracle DSs (Table \ref{tab:Error-DS}). Hence, we treat evidence F1 as a secondary metric to analyze the retrieval performance. Assuming the availability of only IN Set DSs, we compute evidence F1 only for the \textbf{IN Set}, by considering top-K retrieved DSs per query, where K is the number of ground-truth DSs for that query.


\begin{table*}[]
\centering
\caption{Execution Accuracy averaged over ALL \textbf{IN Set} of Dev Set DBs of BirdSQL; NLQ: NL Query; NL: NL Domain Statements (DS); S$\mathrm{tr}$: DBA Screened Structured DS;
QS, No DS, All-DS, BM25, BE, MS-M, STS$\mathrm{b}$, $\mathrm{s}$BSR : Baselines (\S\ref{sec:base}); S$\mathrm{b}$R: Sub-string based Retrieval (\S\ref{sec:approach}); Retrievals with K = 4; \textbf{Bold}: Best performance.}
\footnotesize
\resizebox{\textwidth}{!}{%
\begin{tabular}{l||l||lll|llllllll|llll}
\hline
\multirow{3}{*}{\textbf{LLM}} &
  \multicolumn{1}{c||}{\multirow{3}{*}{\textbf{QS}}} &
  \multicolumn{3}{c|}{\textbf{No Retrieval}} &
  \multicolumn{8}{c|}{\textbf{Retrieval: \emph{Whole} NLQ}} &
  \multicolumn{4}{l}{\textbf{\begin{tabular}[c]{@{}l@{}}\textbf{Retrieval: \emph{Decomposed} NLQ}\end{tabular}}} \\ \cline{3-17} 
 &
  \multicolumn{1}{c||}{} &
  \multicolumn{1}{l|}{\textbf{No}} &
  \multicolumn{2}{c|}{\textbf{All DS}} &
  \multicolumn{2}{c|}{\textbf{BM25}} &
  \multicolumn{2}{c|}{\textbf{BE}} &
  \multicolumn{2}{c|}{\textbf{MS-M}} &
  \multicolumn{2}{c|}{\textbf{STSb}} &
  \multicolumn{2}{c|}{\textbf{sBSR}} &
  \multicolumn{2}{c}{\textbf{SbR}} \\ \cline{4-17} 
 &
  \multicolumn{1}{c||}{} &
  \multicolumn{1}{l|}{\textbf{DS}} &
  \multicolumn{1}{c|}{\textbf{NL}} &
  \multicolumn{1}{c|}{\textbf{Str}} &
  \multicolumn{1}{c|}{\textbf{NL}} &
  \multicolumn{1}{c|}{\textbf{Str}} &
  \multicolumn{1}{c|}{\textbf{NL}} &
  \multicolumn{1}{c|}{\textbf{Str}} &
  \multicolumn{1}{c|}{\textbf{NL}} &
  \multicolumn{1}{c|}{\textbf{Str}} &
  \multicolumn{1}{c|}{\textbf{NL}} &
  \multicolumn{1}{c|}{\textbf{Str}} &
  \multicolumn{1}{c|}{\textbf{NL}} &
  \multicolumn{1}{c|}{\textbf{Str}} &
  \multicolumn{1}{c|}{\textbf{NL}} &
  \textbf{Str} \\ \hline
\textbf{\begin{tabular}[c]{@{}l@{}}Mixtral (8X7B)\end{tabular}} &
 20.2 &
  \multicolumn{1}{l|}{10.3} &
  \multicolumn{1}{l|}{11.4} &
  13.1 &
  \multicolumn{1}{l|}{12.5} &
  \multicolumn{1}{l|}{13.0} &
  \multicolumn{1}{l|}{13.0} &
  \multicolumn{1}{l|}{15.6} &
  \multicolumn{1}{l|}{11.9} &
  \multicolumn{1}{l|}{13.5} &
  \multicolumn{1}{l|}{13.9} &
  16.4 &
  \multicolumn{1}{l|}{13.9} &
  \multicolumn{1}{l|}{14.3} &
  \multicolumn{1}{l|}{14.6} &
  \textbf{17.2} \\ 
\textbf{\begin{tabular}[c]{@{}l@{}}Llama 3 (70B)\end{tabular}} &
  23.5 &
  \multicolumn{1}{l|}{12.2} &
  \multicolumn{1}{l|}{14.4} &
  17.0 &
  \multicolumn{1}{l|}{16.0} &
  \multicolumn{1}{l|}{16.9} &
  \multicolumn{1}{l|}{16.0} &
  \multicolumn{1}{l|}{21.6} &
  \multicolumn{1}{l|}{15.1} &
  \multicolumn{1}{l|}{18.0} &
  \multicolumn{1}{l|}{19.2} &
  22.3 &
  \multicolumn{1}{l|}{17.2} &
  \multicolumn{1}{l|}{19.3} &
  \multicolumn{1}{l|}{21.1} &
  \textbf{25.2} \\ 
\textbf{\begin{tabular}[c]{@{}l@{}}SQL Coder (8B)\end{tabular}} &
  25.3 &
  \multicolumn{1}{l|}{15.3} &
  \multicolumn{1}{l|}{15.8} &
  18.8 &
  \multicolumn{1}{l|}{18.3} &
  \multicolumn{1}{l|}{19.1} &
  \multicolumn{1}{l|}{18.2} &
  \multicolumn{1}{l|}{23.0} &
  \multicolumn{1}{l|}{17.4} &
  \multicolumn{1}{l|}{19.4} &
  \multicolumn{1}{l|}{20.9} &
  24.4 &
  \multicolumn{1}{l|}{19.4} &
  \multicolumn{1}{l|}{20.6} &
  \multicolumn{1}{l|}{24.2} &
  \textbf{30.1} \\ 
\textbf{\begin{tabular}[c]{@{}l@{}}Gemini 1.5 Flash\end{tabular}} &
 40.8 &
  \multicolumn{1}{l|}{30.4} &
  \multicolumn{1}{l|}{34.1} &
  35.2 &
  \multicolumn{1}{l|}{31.6} &
  \multicolumn{1}{l|}{32.1} &
  \multicolumn{1}{l|}{31.0} &
  \multicolumn{1}{l|}{33.9} &
  \multicolumn{1}{l|}{31.5} &
  \multicolumn{1}{l|}{32.5} &
  \multicolumn{1}{l|}{35.5} &
  36.8 &
  \multicolumn{1}{l|}{32.0} &
  \multicolumn{1}{l|}{33.0} &
  \multicolumn{1}{l|}{35.9} &
 \textbf{39.6} \\ 
\textbf{\begin{tabular}[c]{@{}l@{}}GPT 3.5 Turbo \end{tabular}} &
  42.1 &
  \multicolumn{1}{l|}{32.2} &
  \multicolumn{1}{l|}{35.1} &
  39.0 &
  \multicolumn{1}{l|}{35.3} &
  \multicolumn{1}{l|}{41.2} &
  \multicolumn{1}{l|}{33.0} &
  \multicolumn{1}{l|}{42.3} &
  \multicolumn{1}{l|}{31.6} &
  \multicolumn{1}{l|}{40.8} &
  \multicolumn{1}{l|}{37.5} &
  45.8 &
  \multicolumn{1}{l|}{35.3} &
  \multicolumn{1}{l|}{41.4} &
  \multicolumn{1}{l|}{38.3} &
  \textbf{47.5} \\ \hline 
\end{tabular}%
}
\label{tab:main_results_IN}
\end{table*}

\subsection{Baselines}\label{sec:base}
We compare the following methods:
\begin{enumerate*}[label=(\arabic*)]
    \item \textbf{No Retrieval:}
        \begin{enumerate*}[label=(\alph*)]
            \item \textbf{QS}: Query-Specific oracle DSs provided in the dataset, serve as an upper-bound for the OUT set,
            \item \textbf{No-DS}: zero-shot with no DSs,
            \item \textbf{All-DS-NL}: All NL Domain Statements ($\{d_{NL}\}$) provided by domain experts for the IN Set,
            \item \textbf{All-DS-Str}:All DBA screened Structured Domain Statements ($\{d\}$) for the IN set.
        \end{enumerate*}
    \item \textbf{Retrieval:} Retrieval of top-K  NL DSs ($\{\hat{d}_{NL_t}\} \subseteq \{d_{NL}\}$), where similarity is computed with the complete statement $d_{NL}$ and LLM generated Structured (L-Str) and DBA screened Structured (Str) DSs ($\{\hat{d}_{t}\} \subseteq \{d\}$), where the similarity is computed with the `text' ($d.x$) part of $d$.
        \begin{enumerate*}[label=(\alph*)]
        \item \textbf{Similarity with \emph{Whole} NL query $t$}: 
            \begin{enumerate*}[label=(\roman*)]
            \item \textbf{BM25}:
            Okapi variant \cite{robertson1994okapi} of BM25 from the $rank\_bm25$\footnote{\scriptsize \url{https://github.com/dorianbrown/rank_bm25}} library,
            \item \textbf{BE}: 
            cosine similarity using BERT embedding,
            \item \textbf{MS-M}: on the lines of \cite{dou2023towards} we use the best performing\footnote{\scriptsize ckpt ms-marco-MiniLM-L-12-v2: \url{https://www.sbert.net/docs/pretrained-models/ce-msmarco.html}} dense retriever trained with MS MARCO dataset \cite{lei-etal-2023-unsupervised},
            \item \textbf{STSb}: the best performing\footnote{\scriptsize ckpt stsb-roberta-base: \url{https://sbert.net/docs/cross_encoder/pretrained_models.html}} dense retriever trained with STS dataset \cite{cer-etal-2017-semeval},
            \end{enumerate*}
        \item \textbf{Similarity with \emph{Decomposed} NL query}:
        \textbf{(sBSR)}: The method of set selection with BERT Score Recall \cite{gupta-etal-2023-coverage}. Here, we take the contextual representation of each word in the query seperately and perform retrieval of DSs to cover all words.
    \end{enumerate*}
   
\end{enumerate*}

\begin{table}[]

\caption{ Execution accuracy of \textbf{IN Set} of Dev Set DBs by \textbf{GPT-3.5-Turbo}; NL: NL Domain Statements (DS); L-S$\mathrm{tr}$/ S$\mathrm{tr}$:  LLM generated; DBA Screened Structured DS;
QS, No DS, All-DS: Baselines (\S\ref{sec:base});
S$\mathrm{b}$R: Our Sub-string based Retrieval (\S\ref{sec:approach}); Retrievals with  K=4; \textbf{\underline{Bold Underline}}: Best performance; \textbf{Bold}: Second Best.
}
\label{tab:gpt-in-results}
\centering
\small
\resizebox{\columnwidth}{!}{%
\begin{tabular}{l||l||l|l|l|lll}
\hline
 &
   &
   \multicolumn{1}{c|}{\textbf{No}}&
  \multicolumn{2}{c|}{\textbf{All DS}} &
  \multicolumn{3}{c}{\textbf{SbR}} \\ \cline{4-8} 
\multirow{-2}{*}{\textbf{Database}} &
  \multirow{-2}{*}{\textbf{QS}} &
  
  \multicolumn{1}{c|}{\textbf{DS}} &
  \multicolumn{1}{c|}{\textbf{NL}} &
  \textbf{Str} &
  
  \multicolumn{1}{c|}{\textbf{NL}} &
  \multicolumn{1}{c|}{\textbf{L-Str}} &
  \textbf{Str} \\ \hline
\textbf{Thrombosis Pred.} &
  30.9 &
  18.5 &
  \multicolumn{1}{l|}{20.7} &
  13.4 &
 
 \multicolumn{1}{l|}{{24.6}} &
 \multicolumn{1}{l|}{{\textbf{35.8}}} &
  \multicolumn{1}{l}{{\textbf{\underline{37.0}}}} 
  \\ 
\textbf{California Schools} &
  25.0 &
  11.4 &
  \multicolumn{1}{l|}{11.1} &
  {20.0} &
 
 \multicolumn{1}{l|}{{15.9}} &
 \multicolumn{1}{l|}{\textbf{20.5}} &
  \multicolumn{1}{l}{{\textbf{\underline{29.6}}}}
 
  \\ 
 
\textbf{Card Games} &
  32.6 &
  22.1 &
  \multicolumn{1}{l|}{26.0} &
  {\textbf{32.6}} &

 \multicolumn{1}{l|}{{29.4}} &
 \multicolumn{1}{l|}{\textbf{32.6}} &
  \multicolumn{1}{l}{{ \underline{\textbf{47.4}}}} 

  \\
\textbf{Debit Card Spec.} &

  21.9 &
  15.6 &
  \multicolumn{1}{l|}{30.3} &
 \textbf{33.3} &

 \multicolumn{1}{l|}{{28.1}} &
 \multicolumn{1}{l|}{{{31.2}}} &
  \multicolumn{1}{l}{{\textbf{\underline{34.4}}}} 

  \\ 

\multirow{1}{*}{\textbf{Toxicology}} &

  47.2 &
  41.7 &
  \multicolumn{1}{l|}{31.5} &
  {\textbf{46.6}} &

 \multicolumn{1}{l|}{{45.8}} &
 \multicolumn{1}{l|}{\textbf{\underline{48.6}}} &
  \multicolumn{1}{l}{43.1} 
 \\

\multirow{1}{*}{\textbf{Financial}} &

  43.4 &
  34.0 &
  \multicolumn{1}{l|}{35.2} &
  33.3 &

 \multicolumn{1}{l|}{{32.0}} &
 \multicolumn{1}{l|}{{\textbf{35.8}}} &
  \multicolumn{1}{l}{{\textbf{\underline{41.5}}}} 
  \\ 
\textbf{Codebase Comm.} &

  57.0 &
  47.3 &
  \multicolumn{1}{l|}{47.3} &
 54.3 &

 \multicolumn{1}{l|}{\textbf{55.9}} &
 \multicolumn{1}{l|}{{50.5}} &
  \multicolumn{1}{l}{{\textbf{\underline{59.1}}}} 

  \\ 
\textbf{Euro. Football2} &
  35.9 &
  31.3 &
  \multicolumn{1}{l|}{38.5} &
 33.9 &
 \multicolumn{1}{l|}{{35.9}} &
 \multicolumn{1}{l|}{{\textbf{45.3}}} &
  \multicolumn{1}{l}{{\textbf{\underline{53.1}}}} 
 \\ 

\multirow{1}{*}{\textbf{Formula 1}} &
  31.0 &
  28.7 &
  \multicolumn{1}{l|}{27.6} &
  33.0&
 \multicolumn{1}{l|}{{27.5}} &

  \multicolumn{1}{l|}{\textbf{{33.3}}} &
  \multicolumn{1}{l}{{\textbf{\underline{36.8}}}} 
\\
\textbf{Student Club} &
  50.6 &
  39.2 &
  \multicolumn{1}{l|}{40.0} &
  {46.3} &

 \multicolumn{1}{l|}{{48.1}} &
 \multicolumn{1}{l|}{\textbf{\underline{55.6}}} &
  \multicolumn{1}{l}{\textbf{{53.2}}} 
  \\
\multirow{1}{*}{\textbf{Superhero}} &
  75.0 &
  50.0 &
  \multicolumn{1}{l|}{72.3} &
  {75.4} &
 \multicolumn{1}{l|}{{65.6}} &
 \multicolumn{1}{l|}{\textbf{\underline{79.6}}} &
  \multicolumn{1}{l}{{\textbf{76.6}}} 
  \\
 
\hline 
 \textbf{Average} &

  42.1 &
  32.2 &
  \multicolumn{1}{l|}{35.1} &
  39.0 &

 \multicolumn{1}{l|}{{38.3}} &
 \multicolumn{1}{l|}{{\textbf{43.6}}} &
  \multicolumn{1}{l}{{\underline{\textbf{47.5}}} }
  
  \\ \hline
\end{tabular}%
}

\end{table}

\begin{table*}[!ht]
\centering
\caption{Execution Accuracy averaged over ALL \textbf{OUT Set} of Dev Set DBs of BirdSQL; 
Refer Table \ref{tab:main_results_IN} for abbreviations;
QS:\textbf{Upper-bound}; Retrievals with  K = 10.}
\footnotesize
\resizebox{\textwidth}{!}{%
\begin{tabular}{l||l||lll|llllllll|llll}
\hline
\multirow{3}{*}{\textbf{LLM}} &
  \multicolumn{1}{c||}{\multirow{3}{*}{\textbf{QS}}} &
  \multicolumn{3}{c|}{\textbf{No Retrieval}} &
  \multicolumn{8}{c|}{\textbf{ Retrieval: \emph{Whole} NLQ}} &
  \multicolumn{4}{l}{\textbf{\begin{tabular}[c]{@{}l@{}}Retrieval: \emph{Decomposed} NLQ\end{tabular}}} \\ \cline{3-17} 
 &
  \multicolumn{1}{c||}{} &
  \multicolumn{1}{l|}{\textbf{No}} &
  \multicolumn{2}{c|}{\textbf{All DS}} &
  \multicolumn{2}{c|}{\textbf{BM25}} &
  \multicolumn{2}{c|}{\textbf{BE}} &
  \multicolumn{2}{c|}{\textbf{MS-M}} &
  \multicolumn{2}{c|}{\textbf{STSb}} &
  \multicolumn{2}{c|}{\textbf{sBSR}} &
  \multicolumn{2}{c}{\textbf{SbR}} \\ \cline{4-17} 
 &
  \multicolumn{1}{c||}{} &
  \multicolumn{1}{l|}{\textbf{DS}} &
  \multicolumn{1}{c|}{\textbf{NL}} &
  \multicolumn{1}{c|}{\textbf{Str}} &
  \multicolumn{1}{c|}{\textbf{NL}} &
  \multicolumn{1}{c|}{\textbf{Str}} &
  \multicolumn{1}{c|}{\textbf{NL}} &
  \multicolumn{1}{c|}{\textbf{Str}} &
  \multicolumn{1}{c|}{\textbf{NL}} &
  \multicolumn{1}{c|}{\textbf{Str}} &
  \multicolumn{1}{c|}{\textbf{NL}} &
  \multicolumn{1}{c|}{\textbf{Str}} &
  \multicolumn{1}{c|}{\textbf{NL}} &
  \multicolumn{1}{c|}{\textbf{Str}} &
  \multicolumn{1}{c|}{\textbf{NL}} &
  \textbf{Str} \\ \hline 
\textbf{\begin{tabular}[c]{@{}l@{}}Mixtral (8X7B)\end{tabular}} &
  15.9 &
  \multicolumn{1}{l|}{8.0} &
  \multicolumn{1}{l|}{9.1} &
  11.2 &
  \multicolumn{1}{l|}{10.0} &
  \multicolumn{1}{l|}{10.5} &
  \multicolumn{1}{l|}{9.9} &
  \multicolumn{1}{l|}{13.0} &
  \multicolumn{1}{l|}{09.5} &
  \multicolumn{1}{l|}{11.0} &
  \multicolumn{1}{l|}{11.1} &
  13.0 &
  \multicolumn{1}{l|}{10.5} &
  \multicolumn{1}{l|}{11.4} &
  \multicolumn{1}{l|}{12.1} &
  \textbf{14.7} \\
\textbf{\begin{tabular}[c]{@{}l@{}}Llama 3 (70B)\end{tabular}} &
  23.3 &
  \multicolumn{1}{l|}{10.9} &
  \multicolumn{1}{l|}{13.5} &
  15.4 &
  \multicolumn{1}{l|}{14.6} &
  \multicolumn{1}{l|}{15.3} &
  \multicolumn{1}{l|}{17.5} &
  \multicolumn{1}{l|}{19.6} &
  \multicolumn{1}{l|}{13.8} &
  \multicolumn{1}{l|}{16.6} &
  \multicolumn{1}{l|}{19.3} &
  21.1 &
  \multicolumn{1}{l|}{15.4} &
  \multicolumn{1}{l|}{17.0} &
  \multicolumn{1}{l|}{20.5} &
  \textbf{25.6} \\ 
\textbf{\begin{tabular}[c]{@{}l@{}}SQL Coder (8B)\end{tabular}} &
  24.0 &
  \multicolumn{1}{l|}{13.0} &
  \multicolumn{1}{l|}{14.6} &
  16.7 &
  \multicolumn{1}{l|}{16.2} &
  \multicolumn{1}{l|}{17.0} &
  \multicolumn{1}{l|}{18.3} &
  \multicolumn{1}{l|}{21.0} &
  \multicolumn{1}{l|}{15.4} &
  \multicolumn{1}{l|}{17.8} &
  \multicolumn{1}{l|}{19.3} &
  23.2 &
  \multicolumn{1}{l|}{17.9} &
  \multicolumn{1}{l|}{18.8} &
  \multicolumn{1}{l|}{22.2} &
  \textbf{26.0} \\
\textbf{\begin{tabular}[c]{@{}l@{}}Gemini 1.5 Flash\end{tabular}} &
  40.3 &
  \multicolumn{1}{l|}{28.2} &
  \multicolumn{1}{l|}{34.3} &
  32.1 &
  \multicolumn{1}{l|}{29.9} &
  \multicolumn{1}{l|}{30.6} &
  \multicolumn{1}{l|}{29.2} &
  \multicolumn{1}{l|}{33.0} &
  \multicolumn{1}{l|}{30.4} &
  \multicolumn{1}{l|}{31.1} &
  \multicolumn{1}{l|}{33.8} &
  35.4 &
  \multicolumn{1}{l|}{30.7} &
  \multicolumn{1}{l|}{32.1} &
  \multicolumn{1}{l|}{34.6} &
  \textbf{38.7} \\ 
\textbf{\begin{tabular}[c]{@{}l@{}}GPT 3.5 Turbo \end{tabular}} &
  41.2 &
  \multicolumn{1}{l|}{29.7} &
  \multicolumn{1}{l|}{36.6} &
  34.2 &
  \multicolumn{1}{l|}{28.9} &
  \multicolumn{1}{l|}{29.2} &
  \multicolumn{1}{l|}{30.1} &
  \multicolumn{1}{l|}{36.1} &
  \multicolumn{1}{l|}{34.2} &
  \multicolumn{1}{l|}{32.3} &
  \multicolumn{1}{l|}{33.3} &
  32.9 &
  \multicolumn{1}{l|}{35.1} &
  \multicolumn{1}{l|}{34.7} &
  \multicolumn{1}{l|}{34.8} &
  \textbf{39.5} \\ \hline
\end{tabular}%
}

\label{tab:main_results_OUT}
\end{table*}

\begin{table}[!ht]
\centering
\caption{Execution accuracy of \textbf{OUT Set} of Dev Set DBs BirdSQL by \textbf{GPT-3.5-Turbo}; Refer Table \ref{tab:gpt-in-results} for abbreviations; QS: \textbf{Upper-bound}; Retrievals with K=10. 
}
\footnotesize
\resizebox{\columnwidth}{!}{%
\begin{tabular}{l||l||l|l|l|lll}
\hline
 &
   &
   \multicolumn{1}{c|}{\textbf{No}}&
  \multicolumn{2}{c|}{\textbf{All DS}} &
  \multicolumn{3}{c}{\textbf{SbR}} \\ \cline{4-8} 
\multirow{-2}{*}{\textbf{Database}} &

  \multirow{-2}{*}{\textbf{QS}} &
  
  \multicolumn{1}{c|}{\textbf{DS}} &
  \multicolumn{1}{c|}{\textbf{NL}} &
  \textbf{Str} &
  
  \multicolumn{1}{c|}{\textbf{NL}} &
  \multicolumn{1}{c|}{\textbf{L-Str}} &
  \textbf{Str} \\ \hline
 \textbf{Thrombosis Pred.} &

  32.9 &
  {20.7} &
  \multicolumn{1}{l|}{19.3} &
  10.8 &
 \multicolumn{1}{l|}{{\textbf{24.3}}} &
 \multicolumn{1}{l|}{{\textbf{\underline{25.6}}}} &
  \multicolumn{1}{l}{{\textbf{\underline{25.6}}}} 
  \\ 
  \textbf{California Schools} &
  24.4 &
  11.1 &
  \multicolumn{1}{l|}{13.0} &
  {\underline{\textbf{21.7}}} &
 \multicolumn{1}{l|}{{\textbf{{20.0}}}} &
 \multicolumn{1}{l|}{{13.3}} &
  \multicolumn{1}{l}{{\textbf{{20.0}}}} 
  \\ 
 \textbf{Card Games}&
  27.5 &
  14.6 &
  \multicolumn{1}{l|}{19.6} &
 \textbf{20.6} &
 \multicolumn{1}{l|}{{17.7}} &
 \multicolumn{1}{l|}{{19.7}} &
  \multicolumn{1}{l}{{\textbf{\underline{26.0}}}}
  \\ 
 \textbf{Debit Card Spec.}&
  37.5 &
  15.6 &
  \multicolumn{1}{l|}{27.3} &
 {\textbf{33.3}} &
 \multicolumn{1}{l|}{{{28.1}}} &
 \multicolumn{1}{l|}{15.6} &
  \multicolumn{1}{l}{\textbf{\underline{34.4}}} 
  \\ 
 \textbf{Toxicology} &
  32.9 &
  27.0 &
  \multicolumn{1}{l|}{36.5} &
 {\textbf{\underline{40.5}}} &
 \multicolumn{1}{l|}{{{36.9}}} &
 \multicolumn{1}{l|}{35.6} &
  \multicolumn{1}{l}{{{\textbf{39.7}}}} 
  \\ 
 \textbf{Financial} &
  37.7 &

  \textbf{26.4} &
  \multicolumn{1}{l|}{25.9} &
 22.2 &
 \multicolumn{1}{l|}{{22.6}} &
 \multicolumn{1}{l|}{{23.5}} &
  \multicolumn{1}{l}{\textbf{\underline{30.2}}} 
  \\ 
 \textbf{Codebase Comm.}&
  55.9 &
  45.2 &
  \multicolumn{1}{l|}{\textbf{\underline{48.4}}} &
 45.7 &
 \multicolumn{1}{l|}{{46.2}} &

\multicolumn{1}{l|}{45.1} &
  \multicolumn{1}{l}{\textbf{47.3}} 
  \\ 
 \textbf{Euro. Football2}&

  43.0 &
  35.4 &
  \multicolumn{1}{l|}{\textbf{{43.9}}} &
 28.8 &
 \multicolumn{1}{l|}{{35.3}} &
 \multicolumn{1}{l|}{35.3} &
  \multicolumn{1}{l}{{\textbf{\underline{46.2}}}} 
  \\ 

 \textbf{Formula 1} &
  36.8 &
  {35.6} &
  \multicolumn{1}{l|}{{{{\textbf{36.8}}}}} &
  {\underline{\textbf{37.5}}} &
 \multicolumn{1}{l|}{{28.7}} &
 \multicolumn{1}{l|}{31.0} &
  \multicolumn{1}{l}{{{35.6}}} 
  \\ 
\textbf{Student Club} &
  49.4 &
  43.0 &
  \multicolumn{1}{l|}{\textbf{{55.0}}} &
  48.8 &
 \multicolumn{1}{l|}{{48.1}} &
 \multicolumn{1}{l|}{43.0} &
  \multicolumn{1}{l}{{\textbf{\underline{55.7}}}} 
  \\ 
 \textbf{Superhero} &
  70.8 &
  38.5 &
  \multicolumn{1}{l|}{66.7} &
  62.1 &
 \multicolumn{1}{l|}{{\textbf{\underline{69.2}}}} &
 \multicolumn{1}{l|}{52.3} &
  \multicolumn{1}{l}{\textbf{67.7}} 
\\ 
  \hline
  \multirow{1}{*}{\textbf{Average}}&
  41.2 &
  29.7 &
  \multicolumn{1}{l|}{\textbf{36.6}} &
  34.2 &
 \multicolumn{1}{l|}{{34.8}} &
 \multicolumn{1}{l|}{32.4} &
  \multicolumn{1}{l}{\underline{\textbf{39.5}} }

  \\ \hline
\end{tabular}%
}
\label{tab:gpt-out-results}
\end{table}

\begin{table}[t]
\centering
\footnotesize
\caption{Execution Accuracy of GPT-3.5-Turbo with variations in IN-OUT Sets. \textbf{DS}: Number of Union of IN Set Domain Statements averaged over Databases.}
\small
\begin{tabular}{cllll}
\hline

 \textbf{\% IN-OUT Set}
   &
  \multicolumn{1}{c}{\textbf{50-50}} &
  \multicolumn{1}{c}{\textbf{40-60}} &
  \multicolumn{1}{c}{\textbf{30-70}}  &
 \multicolumn{1}{c}{\textbf{20-80}}
  \\ \hline 
 \textbf{DS}
  &
  \multicolumn{1}{c}{128.53} &
  \multicolumn{1}{c}{108.63} &
  \multicolumn{1}{c}{77.00} &
 \multicolumn{1}{c}{53.54} 
  \\ \hline 
IN & 
\multicolumn{1}{l}{47.51} &
\multicolumn{1}{l}{44.65} & 
\multicolumn{1}{l}{46.72} &
\multicolumn{1}{l}{46.81}
\\
  
  OUT &
  \multicolumn{1}{l}{39.48} &
  \multicolumn{1}{l}{33.18} &
   \multicolumn{1}{l}{32.57} &
 \multicolumn{1}{l}{34.43}
 \\ \hline
 
  All &
  \multicolumn{1}{l}{43.48} &
  \multicolumn{1}{l}{37.73} &
   \multicolumn{1}{l}{36.76} &
 \multicolumn{1}{l}{36.86}
 \\ 
 \hline

\end{tabular}%
\label{tab:in-out-ratio}
\end{table}

\begin{table*}[!ht]
\centering
\footnotesize
\caption{Examples of Error Analysis. GT:  Oracle DSs. The retrieved DSs, matching with the oracle DSs are coded with same color. Parts of generated SQLs are color coded in \textcolor{brown}{brown} and \textcolor{red}{red} to indicate the parts that are correct and incorrect, respectively.}
\resizebox{\textwidth}{!}{%
\begin{tabular}{|l|l|l|}
\hline
\textbf{Approach} & \textbf{Structured Domain Statements} & \textbf{SQL} \\ \hline
\multicolumn{3}{|l|}{\textbf{Example 1} \hspace{15pt} \textbf{DB:} Toxicology.\hspace{15pt}{\textbf{Question:} Calculate the average number of oxygen atoms in single-bonded molecules.(Toxic IID Br vs SbR)}} \\ \hline
GT &
\makecell[l]{1. \textcolor{blue}{`single-bonded molecules' refers to}  \textcolor{blue}{bond\_type = `-'} \\ 2. \textcolor{teal}{average number of oxygen atom} 
 \textcolor{teal}{= AVG(element = `o')}} &
\makecell[l]{SELECT CAST(COUNT(DISTINCT CASE WHEN \\ \textcolor{brown}{T1.element = 'o'} THEN T1. molecule\_id ELSE NULL END)\\ AS REAL) / COUNT(DISTINCT T1.molecule\_id)\textcolor{brown}{FROM} \\\textcolor{brown}{atom AS T1 INNER JOIN bond AS T2} ON T1.molecule\_id \\= T2.molecule\_id WHERE \textcolor{brown}{T2.bond\_type = `-'} } \\ \hline

BE-Str &
\makecell[l]{1. \textcolor{teal}{`average number of oxygen atoms in single-bonded} \\\textcolor{teal}{ molecules' refers} \textcolor{teal}{to CAST(COUNT(DISTINCT CASE WHEN} \\\textcolor{teal}{ atom.element = `o' THEN} \textcolor{teal}{atom.molecule\_id ELSE NULL} \\ \textcolor{teal}{END) AS REAL) / COUNT(DISTINCT atom.molecule\_id)} \\ 2. \textcolor{red}{`average number of bonds the atoms have'} refers to \textcolor{red}{CAST(}\\\textcolor{red}{ COUNT(connected.bond\_id) AS REAL) / COUNT(atom.atom\_id)} \\ 
3. \textcolor{red}{`percentage of carbon in double-bond molecules' refers to} \textcolor{red}{CAST}\\ \textcolor{red}{(COUNT(DISTINCT CASE WHEN atom.element} = \textcolor{red}{'c' THEN}\\ \textcolor{red}{atom.atom\_id ELSE NULL END) AS REAL)}\\ \textcolor{red}{* 100 / COUNT} \textcolor{red}{(DISTINCT atom.atom\_id}} &
\makecell[l]{SELECT CAST(COUNT(DISTINCT CASE WHEN  \\ \textcolor{brown}{atom.element = 'o' AND bond.bond\_type = `-'} THEN  \\ atom.molecule\_id  ELSE NULL END) AS REAL) / \\ COUNT (DISTINCT  atom.molecule\_id) AS \\ average\_oxygen\_atoms FROM \textcolor{brown}{atom JOIN bond} ON \\ atom.molecule\_id =  bond.molecule\_id WHERE \\ \textcolor{brown}{bond.bond\_type = `-'};} \\ \hline

SbR-Str &
\makecell[l]{1. \textcolor{teal}{`average number of oxygen atoms in single-bonded molecules'} \\ \textcolor{teal}{refers} \textcolor{teal}{to CAST(COUNT(DISTINCT CASE WHEN atom.element} \\ \textcolor{teal}{ = 'o' THEN} \textcolor{teal}{atom.molecule\_id ELSE NULL END) AS REAL)} \\ \textcolor{teal}{ / COUNT(DISTINCT atom.molecule\_id)} \\ 2. \textcolor{blue}{`single-bonded molecules' refers to bond.bond\_type = `-'} \\ 3. \textcolor{teal}{`oxygen' refers to atom.element = `o'}} &
\makecell[l]{SELECT CAST(COUNT(DISTINCT CASE WHEN \\ \textcolor{brown}{atom.element = `o'} THEN atom.molecule\_id ELSE NULL \\END) AS REAL) / COUNT\ (DISTINCT atom.molecule\_id) \\FROM \textcolor{red}{atom JOIN molecule ON} \textcolor{red}{ atom.molecule\_id = molecule.} \\ \textcolor{red}{molecule\_id JOIN bond ON} atom.molecule\_id = bond.molecule \\ \_id  WHERE \textcolor{brown}{bond.bond\_type = `-'} AND \textcolor{brown}{atom.element = `o'};} \\ \hline

\multicolumn{3}{|l|}{\textbf{Example 2} \hspace{15pt}  \textbf{DB:} Formula 1.\hspace{15pt}{\textbf{Question:} List circuits which host 4 f1 races from year 990 to 2000.}} \\ \hline

GT &
\makecell[l]{1. \textcolor{teal}{`from year 1990 to 2000 refers to year(date)} \\ \textcolor{teal}{between 1990 and 2000'}\\ 2. \textcolor{blue}{`did not host any f1 races refers to COUNT} \textcolor{blue}{(raceId) = 0'}} &
\makecell[l]{SELECT DISTINCT T1.name FROM circuits AS T1 INNER  \\ JOIN races AS T2 ON T2.circuitID = T1.circuitId WHERE \\ \textcolor{brown}{STRFTIME('\%Y', T2.date) BETWEEN '1990' AND '2000'} \\ GROUP BY T1.name HAVING  \textcolor{brown}{ COUNT(T2.raceId) = 4}} \\ \hline
SbR-Str &
\makecell[l]{1. \textcolor{red}{`In which Formula\_1 race' refers to }\textcolor{red}{DISTINCT races.name} \\2. \textcolor{teal}{`From (100) to (100)' refers to race.year }\\ \textcolor{teal}{BETWEEN 100 AND 100} \\} &
\makecell[l]{
SELECT circuits.name FROM circuits JOIN races ON circuits.\\
circuitId=races.circuitId JOIN seasons ON races.year=seasons.\\year
WHERE \textcolor{brown}{seasons.year} \textcolor{brown}{BETWEEN 1990 AND 2000} GROUP\\
BY circuits.name HAVING \textcolor{red}{COUNT(DISTINCT} \textcolor{red}{races.name) = 4}} \\ \hline

\multicolumn{3}{|l|}{\textbf{Example 3} \hspace{15pt}  \textbf{DB:} Toxicology.\hspace{15pt}{\textbf{Question:} Enumerate the bond ID of triple bond type molecules and tell me if they are carcinogenic or not.}} \\ \hline

GT &
\makecell[l]{
1. \textcolor{teal}{triple bond refers to bond\_type = `\#'} \\ 2. \textcolor{blue}{label = '+' mean molecules are carcinogenic} \\ 3. \textcolor{red}{label = '-' means molecules are non-carcinogenic}} &
\makecell[l]{ \textcolor{brown}{SELECT DISTINCT T1.bond\_id} , T2.label FROM  bond \\  AS T1 INNER JOIN molecule  AS T2 ON T1.molecule\_id \\ = T2.molecule\_id WHERE T1.bond\_type = `\#'} \\ \hline

All-DS-Str&
\makecell[l]{ Not Applicable } &
\makecell[l]{\textcolor{red}{SELECT bond.bond\_id}, molecule.label FROM bond JOIN \\   molecule ON bond.molecule\_id =  molecule.molecule\_id \\ WHERE bond.bond\_type = `\#';} \\ \hline

SbR-Str &
\makecell[l]{ 1. \textcolor{teal}{`triple bond' refers to bond.bond\_type = `\#'} \\ 2. \textcolor{blue}{'carcinogenic' refers to molecule.label = `+'} \\ 3. \textcolor{teal}{`molecules with a triple bond type' refers to } \textcolor{teal}{bond.bond\_type = '\#'} \\
4. \textcolor{blue}{`carcinogenic-type molecules' refers to } \\ \textcolor{blue}{molecules with molecule.label = `+'}} &
\makecell[l]{\textcolor{red}{SELECT bond\_id}, molecule.label FROM bond JOIN  molecule \\ ON bond.molecule\_id = molecule.molecule\_id WHERE \\ bond.bond\_type = `\#' AND   \textcolor{red}{molecule.label = `+'};} \\ \hline
\end{tabular}%
}
\label{tab:error_analysis}
\end{table*}

\subsection{Results}\label{sec:results}

We call our Sub-string based Retrieval as \textbf{SbR}.
We answer the following research questions (RQ): 
\\
\textbf{RQ1: How effective are DSs for the NL-to-SQL task?}
Comparing No DS to QS, across LLMs, Sets and DBs (Tables \ref{tab:gpt-in-results} and \ref{tab:gpt-out-results}) we show the benefit for having  DSs relevant to the query in context for SQL generation. 
\\
\textbf{RQ2: Does our retrieval based approach yield comparable performance  to when we have oracle DSs for a query? }
For the IN Set, with availability of the required DSs, our SbR-Str approach performs better than QS, for 9 out of 11 DBs (Table \ref{tab:gpt-in-results}) and 3 out of 5 LLMs (Table \ref{tab:main_results_IN}). For the remaining DBs and LLMs our approach's performance is comparable. The better performance is the result of: (i) DBA rectification of erroneous LLM structured DSs, 
(examples in Table \ref{tab:Error-DS}), 
and (ii) retrieval of non oracle yet relevant top-4 DSs, boosting the performance for queries with missing and insufficient oracle DSs. As expected, for the OUT Set, with only partial availability of required DSs, our SbR-Str performance is inferior to QS for 3 out of 5 LLMs (Table \ref{tab:main_results_OUT}) and 8 out of 11 DBs (Table \ref{tab:gpt-out-results}). Here, QS acts as the upper-bound.  For the OUT-NO Set our best performing GPT-3.5-Turbo model yields 36.6\% accuracy with SbR-Str, 30.1\% with No DS and 41.7\% with QS. With no availability of DSs the drop in the accuracy of SbR-Str over QS for OUT-NO Set is more (5.1\%) as compared to the OUT Set with partial availability of DSs (1.7\%). Thus, the proposed method (SbR-Str) is designed to harness existing DSs in the repository, relevant to user queries (IN set), but  is not intended to aid queries with no relevant DSs in the repository (OUT sets). However, non-trivial performance gains of SbR-Str over QS for the IN set establishes the efficacy of our core novelty in terms of (i) DS Structuring, and (ii) sub-string based retrieval.\\
\textbf{RQ3: Is retrieving relevant domain  statements better than providing all DSs?}
For both the IN and OUT Sets and all LLMs, QS and our SbR-Str performance is consistently better than All-DS/NL and All-DS/Str (Tables \ref{tab:main_results_IN} and \ref{tab:main_results_OUT}). 
For the IN Set,  SbR-Str  performs better than All-DS/NL and ALL-DS/Str across almost all DBs (Table \ref{tab:gpt-in-results}). This shows that even if the prompt size is not a concern, having relevant DSs in-context is important.  Extensive manual inspection for the incorrect samples resulting from the SQLs generated with SbR-Str retrieved DSs in-context, but correct answers with All-DS/Str demonstrates that most of the failure cases fall in one of the three categories: (i) Retrieved DSs with SbR-Str are relevant to the NL query, however the LLM generates incorrect SQLs with these correct DSs in the context. For e.g., the SQL generated for Example 3 (Table \ref{tab:error_analysis}) with SbR-Str is incorrect though the retrievals are correct. It uses an additional constraint of molecule.label = `+' instead of projecting the labels of the molecule as the part of SELECT clause.  (ii) The generated SQLs with All-DS-Str are semantically wrong, but coincidentally lead to the correct answer. For the same example, the All-DS-Str approach leads to  incorrect SQL without the use of the keyword DISTINCT, but coincidentally gives the same answer as ground truth SQL, (iii) The generated SQLs by SbR-Str are incorrect due to wrong retrievals.  In Example 2(Table \ref{tab:error_analysis}) our SbR-Str approach retrieves an incorrect structured DS due to the high similarity of its `text' part  (`In which formula\_1 race') with the query sub-string (`which host 4 f1 races'). This leads to the generation of incorrect SQL wrongly clued to utilize the  `races.name' from the retrieved DS in the HAVING clause, as opposed to `races.raceId'.

As expected, for the OUT Set, we observe that All-DS/Str and All-DS/NL performs better than SbR-Str for 3 (California School, Toxicology, Formula 1) 
and 2  (Codebase Community, Formula 1) 
out of 11 DBs, respectively. However, For the other DBs, better performance of SbR-Str is due to retrieval of the relevant DSs  with higher recall (K=10), for the queries having partial DS overlap, and missing or in-correct DSs, which get overlooked during overlap computations (\S~\ref{sec:dataset}).
\\
\textbf{RQ4: What is the benefit of structuring DSs?}
For the IN Set, with all LLMs, all the methods perform better with DBA scanned structured (Str) DSs, as opposed to when NL DSs in context (Table \ref{tab:main_results_IN}). 
We observe the same for the OUT Set, except for GPT-3.5-turbo LLM, for 3 retrieval methods (Table \ref{tab:main_results_OUT}). 
For the IN Set,  SbR-L-Str is inferior to SbR-Str due to error propagation because of noisy LLM generated structured DSs. However, for 10 out of 11 DBs, it performs better than SbR-NL (Table \ref{tab:gpt-in-results}).  This demonstrates the effectiveness of LLM based Structuring, allowing complete automation and handling variations in the NL DSs across DBs.
On the other hand, for the OUT Set, we observe inferior performance of SbR-L-Str over SbR-NL for 5 out of 11 DBs. However, after DBA screening, SbR-Str demonstrates better performance than SbR-NL for 10 out of 11 DBs (Table \ref{tab:gpt-out-results}). This demonstrates that the queries with unavailability of all the required DSs, need non-erroneous, DBA scanned structured DSs in context, to achieve  performance comparable to the queries with availability of DSs.
\\
\textbf{RQ5: How important is it to decompose a user query into sub-strings when matching with DSs?} 
Our SbR-Str approach, which retrieves using semantic match with NL query sub-strings  performs better than all the four baselines which use the \emph{whole} query for matching, across LLMs and Sets (Tables \ref{tab:main_results_IN} and \ref{tab:main_results_OUT}). We observe similar trend with Evidence F1 scores with structured DSs (Str) (BM25: 0.35, BE: 0.33, MS-M: 0.34, STSb: 0.37, sBSR: 0.35, SbR: 0.39). At inference time, we can not differentiate between IN and OUT queries. The performance of SbR-Str with DSs retrieved using consistent K = 10 across the sets (42.6), is the best surpassing the baselines (QS: 41.7, NO-DS: 30.9, All-DS-NL: 35.8, All-DS-Str: 36.6,  BM25-Str: 37.3, MSM-Str: 38.4, sBSR-Str: 38.0, BE-Str: 31.4,  STSb-Str: 38.4,  SbR-NL: 36.6 and SbR-L-Str: 36.5). This shows the importance of query decomposition. Superior performance of  SbR-Str approach to sBSR, treating each word of the query as its decomposition, demonstrates efficacy of our sub-strings based query decomposition strategy.

For  IN Sets of 4 DBs (Debit Card Specializing, Toxicology, Formula 1 and Student Club) and OUT Sets of 3 DBs (Card Games, Codebase Community and Formula 1) one of the other retrieval baselines (mostly BE-Str) performs better than SbR-Str. We perform an extensive analysis  for the samples for which we do not get correct answers from the SQLs generated using our approach (SbR-Str).  We find that the errors belongs to the 3 categories explained as the part of discussion of \textbf{RQ3}. For e.g.,  the structured DSs retrieved with our SbR-Str based approach  of the Example 1(Table \ref{tab:error_analysis}), match with the oracle DSs, yet the SQL generated has a unnecessary JOIN on `molecule' table, which makes the SQL query incorrect. Note that the retrieval done using SbR-Str is better than that with the embedding base (BE-Str) approach, where some of the retrieved DSs do not match with oracle DSs. However, in this case, the SQL generated with BE-Str is semantically wrong but coincidentally leads to the correct answer. \\
\textbf{RQ6: How does the size of the DS repository affects the performance.}
As discussed in Section \ref{sec:dataset}, we assume 50-50 IN-OUT \% splits. However, to check the effect of availability of distinct size of the domain repository,  we create distinct IN-OUT \% splits. 
 Table \ref{tab:in-out-ratio} demonstrates the performance of our SbR-Str approach with distinct IN-OUT Sets with the best performing GPT-3.5-Turbo LLM. The number of DSs in the repository of a DB decreases with the decreases in the \% IN Set. As expected this negatively affects the overall performance, especially due to decrease in the performance of the OUT Set, with fewer available DS in the repository. However, IN Set performance  remains consistent, due to availability of DSs for all the queries in that set. This also underscores the effectiveness of DSs for the NL-to-SQL task. 

\section{Conclusion}

This paper presents a framework for expressing, representing, indexing, and retrieving domain knowledge required to convert domain-specific NL questions to SQL.  For enterprise databases, it provides a significantly more practical approach of providing database-level domain statements, than existing NL hints attached to each query.
Results on a dataset derived from BirdSQL demonstrates the value of the domain statement construction, structuring \& retrieval based augmentation for downstream SQL generation. 
Our  novel retrieval strategy outperforms all the baseline approaches.  In future, for new NL user queries we plan to automatically identify unresolved domain-specific expressions and route to domain experts for disambiguation. Finally, to handle enterprise DBs that get frequently updated, we plan to extend our pipeline to accommodate real-time updates of the domain repository.

\bibliographystyle{./IEEEtran}
\bibliography{./main}

\end{document}